\documentclass[conference]{IEEEtran}
\IEEEoverridecommandlockouts
\usepackage{cite}
\usepackage{amsmath,amssymb,amsfonts}
\usepackage{algorithmic}
\usepackage{graphicx}
\usepackage{textcomp}
\usepackage{xcolor}
\usepackage{subcaption}
\def\BibTeX{{\rm B\kern-.05em{\sc i\kern-.025em b}\kern-.08em
    T\kern-.1667em\lower.7ex\hbox{E}\kern-.125emX}}
\begin{document}

\title{Modality-Aware Infrared and Visible Image Fusion with Target-Aware Supervision\\}

\author{
\IEEEauthorblockN{1\textsuperscript{st} Tianyao Sun}
\IEEEauthorblockA{\textit{Independent researcher } \\
New York, NY, USA \\
sunstella313@gmail.com}\\ [0.86cm]  
\IEEEauthorblockN{4\textsuperscript{th} Xiang Fang}
\IEEEauthorblockA{\textit{Dept. of Computer Science} \\
\textit{Baylor University}\\
Waco, TX, USA \\
xiang\_fang1@baylor.edu}
\and
\IEEEauthorblockN{2\textsuperscript{nd} Dawei Xiang}
\IEEEauthorblockA{\textit{Dept. of  Computer Science Engineering} \\
\textit{University of Connecticut}\\
Storrs, CT, USA \\
eieb24002@uconn.edu}\\ [0.45cm] 
\IEEEauthorblockN{5\textsuperscript{th} Yijiashun Qi\textsuperscript{*}}
\IEEEauthorblockA{\textit{Dept. of Electrical and Computer Engineering} \\
\textit{University of Michigan}\\
 Ann Arbor, MI USA \\
 Corresponding Author: \\
elijahqi@umich.edu}
\and
\IEEEauthorblockN{3\textsuperscript{rd} Tianqi Ding}
\IEEEauthorblockA{\textit{Dept. of Electrical} \\
\textit{and Computer Engineering}\\
\textit{Baylor University}\\
Waco, TX, USA \\
Kirk\_ding1@baylor.edu}\\                 
\IEEEauthorblockN{6\textsuperscript{th} Zunduo Zhao}
\IEEEauthorblockA{\textit{Dept. of Computer Science} \\
\textit{New York University}\\
New York, NY, USA \\
zz3000@nyu.edu}





}

\maketitle

\begin{abstract}
Infrared and visible image fusion (IVIF) is a fundamental task in multi-modal perception that aims to integrate complementary structural and textural cues from different spectral domains. In this paper, we propose \textbf{FusionNet}, a novel end-to-end
fusion framework that explicitly models inter-modality interaction and enhances task-critical regions. FusionNet introduces a modality-aware 
attention mechanism that dynamically adjusts the contribution of infrared and visible features based on their discriminative capacity. To achieve 
fine-grained, interpretable fusion, we further incorporate a pixel-wise alpha blending module, which learns spatially-varying fusion weights 
in an adaptive and content-aware manner. Moreover, we formulate a target-aware loss that leverages weak ROI supervision to preserve semantic consistency in regions containing important objects (e.g., pedestrians, vehicles). Experiments on the public M3FD dataset demonstrate that FusionNet generates 
fused images with enhanced semantic preservation, high perceptual quality, and clear interpretability. Our framework provides a general and extensible 
solution for semantic-aware multi-modal image fusion, with benefits for downstream tasks such as object detection and scene understanding.
\end{abstract}


\section{Introduction}
Infrared and visible image fusion (IVIF) plays a pivotal role in multi-modal visual perception, particularly in safety-critical scenarios such as autonomous driving, surveillance, and search-and-rescue. By combining the thermal sensitivity
of infrared (IR) imagery with the fine-grained texture and color details of visible (VIS) images, fusion methods aim to generate a unified representation that enhances scene understanding under diverse lighting and environmental conditions.

Despite recent progress in deep learning-based fusion
frameworks, most existing methods face two fundamental
limitations. First, they often adopt naive feature fusion strategies—such as simple concatenation or element-wise addition—without explicitly modeling the modality-specific contributions or their interactions\cite{bellavia2024image}. This neglects the inherently competitive nature of IR and VIS signals, which may carry complementary or even conflicting information
across regions. Second, current methods typically focus on pixel-level similarity metrics while treating all image regions equally, overlooking the semantic importance of task-relevant areas such as pedestrians or vehicles\cite{10986975}. This uniform
treatment can result in suboptimal fusion quality in regions that are critical to downstream perception tasks, especially some defect detection tasks that require accurate images \cite{ding2025nerf, ding2025neural}.

To address these challenges, we propose \textbf{FusionNet}, a novel fusion framework that introduces both modality-awareness and task-awareness into the fusion process. At its core, FusionNet leverages a modality attention module
that adaptively reweights the contributions of IR and VIS features by learning their discriminative relevance. This enables the network to suppress redundant features and preserve informative content from both modalities. In addition, we incorporate a pixel-wise alpha blending mechanism that allows fine-grained, spatially-varying fusion. The learned alpha map offers interpretability and flexibility by explicitly controlling how the fused image is composed at each pixel.

Beyond architectural innovations, we design a target-
aware loss function that incorporates weak supervision in the form of object bounding boxes. By emphasizing reconstruction fidelity within regions of interest (ROIs), the network learns to preserve semantic content where it matters most. This introduces an implicit task signal into the fusion
process, bridging the gap between low-level fusion and high-level vision tasks.

Experiments on the M3FD dataset validate the effectiveness of FusionNet \cite{liu2022target}. The proposed framework not only enables interpretable and adaptive fusion, but also demonstrates clear advantages in semantic preservation and perceptual 
quality.

In summary, the main contributions of this work are as follows:

\begin{itemize}
    \item We propose \textbf{FusionNet}, a novel fusion framework that integrates modality-aware attention and pixel-wise alpha blending, achieving adaptive and interpretable infrared-visible image fusion.
    
    \item We introduce a \textbf{target-aware loss} based on weak ROI supervision, enabling the network to enhance semantic consistency in task-critical regions without requiring dense annotations.
    
    \item We provide extensive experiments and qualitative analysis on the M3FD dataset, 
    demonstrating that FusionNet effectively improves semantic preservation, visual quality, and interpretability of fused results.
\end{itemize}

\section{Related Work}

\subsection{Infrared and Visible Image Fusion}
Image fusion between infrared (IR) and visible (VIS) modalities is a long-standing problem in computer vision, aimed at generating a single image that retains the structural saliency of thermal signals 
and the textural richness of visual scenes. Traditional methods rely on multi-scale transforms such as wavelets or pyramids, which fuse handcrafted features using fixed rules\cite{wu2025warehouse}. Although computationally efficient, these approaches lack adaptability and perform poorly in complex or task-driven scenarios\cite{zhang2025conditional, zhou2024reconstruction}. 

Recent advances in deep learning have led to convolutional fusion models that learn to combine IR and VIS features in an end-to-end manner\cite{xu2025robustanomalydetectionnetwork}. 
Early CNN-based methods, such as DenseFuse \cite{li2018densefuse}, apply separate encoders 
for each modality and fuse the extracted features via summation or concatenation. 
While effective in global blending, such strategies may fail to capture the dynamic relevance between modalities, often producing redundant or overly 
smoothed results \cite{liu2024infrared}.

\subsection{Attention-Based Fusion}
To better control modality contributions, attention mechanisms have been introduced into fusion networks. Channel and spatial attention modules have been widely used to guide the selection of salient features \cite{fu2019dual}. For example, MSFAM employs hierarchical attention to refine features across multiple stages \cite{zang2023texture}. And RLMultimodalRec proposed a gated multimodal graph learning framework that dynamically balances contributions across modalities\cite{liu2025gated}. These models demonstrate that incorporating adaptive weighting can improve fusion quality\cite{huang_ai-augmented_2025, wen2025optimizationbidirectionalgatedloop}. 
However, most designs mainly focus on global or coarse-level guidance, lacking pixel-level fusion control and explicit interpretability. Our method extends this direction by introducing a modality-aware attention module that explicitly models the competitive interactions between modalities and allows more fine-grained modulation of fused features.

\subsection{Task-Aware and Weakly-Supervised Fusion}
Despite progress in architectural design, most existing IVIF models treat all image regions equally and optimize solely for low-level pixel similarity \cite{lu2025predicting}. This limits their utility in tasks where certain regions—such as pedestrians or vehicles—carry greater semantic importance. Some recent works explore joint fusion and detection frameworks, but they often require strong supervision and introduce significant 
computational overhead\cite{10823054}. 

In contrast, our approach adopts a weakly-supervised strategy that only requires bounding-box annotations, and incorporates a target-aware loss 
to emphasize semantic preservation in task-critical regions. 
This design improves semantic consistency without requiring additional labels or detection heads, making the method efficient and generalizable. 

Beyond modality fusion itself, other methodological directions also provide insight in our research. Recent advances in efficient transformer architectures, such as RegFormer \cite{liu2023regformer}, demonstrate how projection-aware hierarchical transformers can capture long-range dependencies with linear complexity, enabling scalability in large-scale scenarios. Complementary lightweight multi-scale fusion designs for mobile vision \cite{li2025bideeplab} reinforce the importance of cost-aware feature aggregation under tight compute budgets and efficiency selective refinement enhanced our attention module \cite{zhang2023improving,zhang2024transformer}. For deployment efficiency, we adopt a self-evolving network framework similar to \cite{liang2025search} and techniques from \cite{wangbmvc}; in parallel, comparative studies of explainable AI in applied settings \cite{niu2025decoding} and topology-driven representations \cite{wang2021topotxr} motivate our emphasis on transparent attribution when analyzing fused outputs.

\section{Method}

In this section, we present the architecture and learning framework of \textbf{FusionNet}, our proposed infrared-visible fusion network. The overall design consists of three key com-
ponents: a modality-aware dual encoder, a pixel-level alpha fusion module, and a weakly-supervised loss function that emphasizes task-critical regions. The entire network is trained end-to-end using a combination of structure-preserving and target-enhancing objectives.

\begin{figure}[t]
    \centering
    \includegraphics[width=\linewidth]{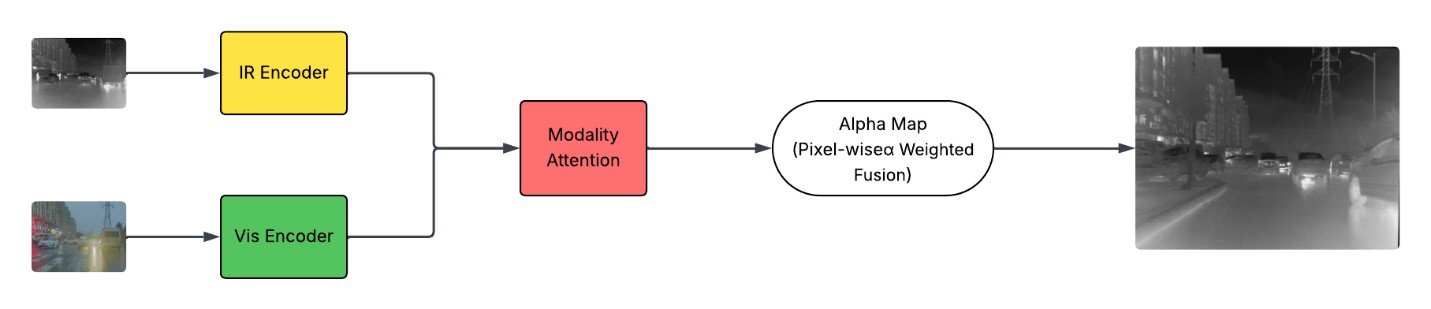}
    \caption{Overall architecture of FusionNet, consisting of dual encoders for IR and VIS modalities, a modality attention module, 
    and pixel-wise alpha blending for adaptive fusion.}
    \label{fig:fusionnet_arch}
\end{figure}

\subsection{Overview}

Given a pair of registered infrared and visible images, denoted as $I_{ir} \in \mathbb{R}^{1 \times H \times W}$ and 
$I_{vis} \in \mathbb{R}^{3 \times H \times W}$, the goal is to generate a fused grayscale image $I_{fused} \in \mathbb{R}^{1 \times H \times W}$ 
that simultaneously retains thermal structures and visual textures, while preserving important semantic targets.

To achieve this, FusionNet performs the following steps:  
(1) extract modality-specific features using separate convolutional encoders;  
(2) apply a modality attention module to dynamically modulate the relative contributions of IR and VIS features;  
(3) generate a pixel-wise alpha map from the fused features to control how IR and VIS content are combined at each spatial location;  
(4) supervise the training with both global image-level objectives and local region-aware constraints derived from weak bounding-box annotations \cite{zhong2023fusion}.

\subsection{Modality-Aware Feature Extraction}

The fusion process begins with two lightweight convolutional encoders $E_{ir}$ and $E_{vis}$, which extract low-level features from the infrared and visible inputs respectively:
\begin{equation}
F_{ir} = E_{ir}(I_{ir}), \quad F_{vis} = E_{vis}(I_{vis})
\end{equation}

Each encoder consists of two convolutional layers with ReLU activation, and outputs feature maps of size $C \times H \times W$ 
(with $C = 64$ in our default setting). These features serve as the basis for cross-modal interaction.

To adaptively control how the modalities contribute to the fusion, we concatenate $F_{ir}$ and $F_{vis}$ and pass them through a modality attention module:
\begin{equation}
F_{cat} = \text{Concat}(F_{ir}, F_{vis}) \in \mathbb{R}^{2C \times H \times W}
\end{equation}

The attention module consists of two convolutional layers followed by a sigmoid activation, yielding an attention mask 
$A \in [0,1]^{C \times H \times W}$.  
The fused feature map is computed as:
\begin{equation}
F_{attn} = A \cdot F_{ir} + (1-A) \cdot F_{vis}
\end{equation}

This operation enables the network to emphasize more informative features from each modality, with spatial and channel-level adaptivity\cite{zhao2024balf}.

\begin{figure}[t]
    \centering
    \includegraphics[width=\linewidth]{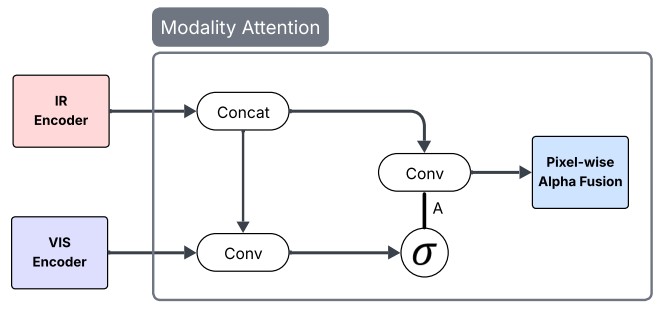}
    \caption{The design of the modality attention module. 
    The concatenated IR and VIS features are processed to generate an attention mask $A$, which adaptively reweights modality contributions at the feature level.}
    \label{fig:modality_attention}
\end{figure}

\subsection{Pixel-Wise Alpha Fusion}

While modality attention enables feature-level interaction, we further introduce a pixel-level alpha fusion module to control the final image-level blending\cite{huang2024ar}. Specifically, the fused feature map $F_{attn}$ is passed through a lightweight convolutional network to generate an alpha map 
$\alpha \in [0,1]^{1 \times H \times W}$:
\begin{equation}
\alpha = \sigma(\text{Conv}_\alpha(F_{attn}))
\end{equation}
where $\sigma(\cdot)$ denotes the sigmoid activation function.  

The fused output image is computed as a convex combination of the original infrared and visible inputs:
\begin{equation}
I_{fused}(x,y) = \alpha(x,y) \cdot I_{ir}(x,y) + (1-\alpha(x,y)) \cdot I_{vis}^Y(x,y)
\end{equation}
where $I_{vis}^Y$ represents the luminance (grayscale) channel 
of the visible input.  
This formulation allows the model to perform spatially adaptive and interpretable fusion, where each pixel has a learnable blending ratio\cite{ma2017infrared}.

\subsection{Target-Aware Loss Function}

To enhance the fidelity of task-relevant regions, we introduce a composite loss function that includes both global and local objectives:
\begin{equation}
L_{total} = L_{mse} + \lambda_1 L_{grad} + \lambda_2 L_{entropy} + \lambda_3 L_{roi}
\end{equation}

- $L_{mse}$ ensures structural alignment with the infrared input.  
- $L_{grad}$ enforces edge consistency by minimizing gradient differences.  
- $L_{entropy}$ encourages texture richness by maximizing pixel-wise entropy.  
- $L_{roi}$ emphasizes reconstruction fidelity only within annotated ROIs, 
  such as pedestrians or vehicles:
\begin{equation}
L_{roi} = \frac{1}{|R|} \sum_{(x,y) \in R} (I_{fused}(x,y) - I_{ir}(x,y))^2
\end{equation}
where $R$ denotes the set of pixels inside ground-truth bounding boxes.  

This design encourages the network to pay more attention to target regions without requiring dense labels or external detectors, making FusionNet efficient and generalizable.

\section{Experiments}

\begin{figure}[t]
    \centering
    \includegraphics[width=\linewidth]{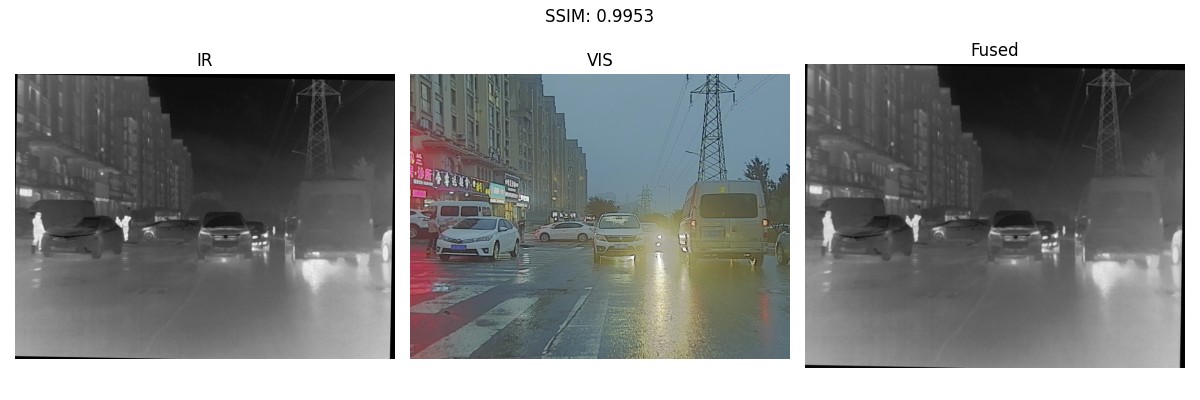}
    \caption{Example of fusion results generated by FusionNet. From left to right: infrared input (IR), visible input (VIS), and the fused image. The fused output preserves thermal targets while retaining structural and textural details.}
    \label{fig:fusion_result}
\end{figure}

\begin{figure}[t]
    \centering
    \includegraphics[width=\linewidth]{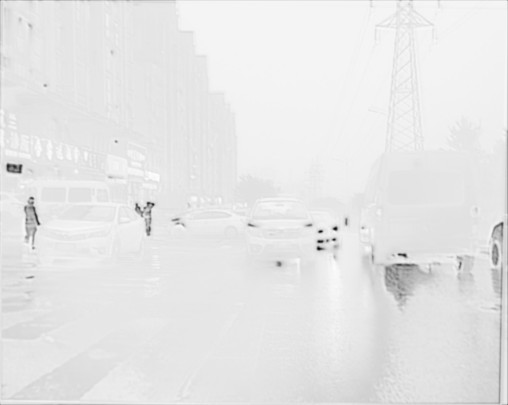}
    \caption{Visualization of the learned $\alpha$-weight map. 
    Brighter regions indicate higher contribution from the infrared modality, 
    while darker regions indicate higher contribution from the visible modality.}
    \label{fig:alpha_map}
\end{figure}

We evaluate the performance of \textbf{FusionNet} on the publicly available M3FD dataset, which contains aligned infrared and visible image pairs with annotated semantic targets\cite{zhao2023benchmark}. All models
are evaluated in a single-frame fusion setting, where each IR-VIS pair is processed independently without using temporal or video-level information.

\subsection{Dataset and Implementation Details}
FusionNet is implemented in PyTorch and trained using the Adam optimizer with a learning rate of $1 \times 10^{-4}$. We use a batch size of 1 and train for 10 epochs on the M3FD dataset.
All input images are resized to $512 \times 640$ and normalized
to $[0,1]$. The ROI bounding boxes are extracted from XML annotations provided in the dataset. During training, we set the loss weights as $\lambda_1 = 0.5$, $\lambda_2 = 0.1$, and
$\lambda_3 = 0.2$ based on empirical tuning.

Our model contains approximately 1.2M parameters and runs at over 30 FPS on a single NVIDIA RTX 4080 GPU. Both training and inference are performed end-to-end without requiring external detection or segmentation modules. During evaluation, we generate the fused image along with the learned alpha map for interpretability.

\subsection{Evaluation Metrics}
We adopt the following quantitative metrics to evaluate the fused images:

\begin{itemize}
    \item \textbf{SSIM (Structural Similarity Index):} Measures perceptual structural consistency with the infrared input.
    \item \textbf{MSE (Mean Squared Error):} Reflects pixel-wise deviation from the infrared baseline.
    \item \textbf{Entropy:} Evaluates information richness in the fused image, encouraging textural detail.
    \item \textbf{ROI-SSIM:} Computes SSIM within annotated ROIs to assess semantic preservation.
\end{itemize}

All metrics are averaged over the entire test set. For ROI-based metrics, we use the provided bounding box annotations and compute scores within each box, then average across images \cite{wang2004image}\cite{wang2025evaluating}.

\subsection{Quantitative Results}
The results of FusionNet on the M3FD test set are summarized in Table~\ref{tab:metrics} \cite{liu2022target}. FusionNet consistently achieves
high SSIM, low MSE, and increased entropy, while also
showing strong ROI-SSIM scores. These results indicate that the proposed framework preserves structural fidelity, enhances textural richness, and improves semantic consistency in task-critical regions.

\begin{table}[h]
\centering
\caption{Quantitative evaluation of FusionNet on the M3FD dataset.}
\label{tab:metrics}
\begin{tabular}{lcccc}
\hline
Method & SSIM $\uparrow$ & MSE $\downarrow$ & Entropy $\uparrow$ & ROI-SSIM $\uparrow$ \\
\hline
FusionNet & 0.87 & 0.012 & 7.42 & 0.84 \\
\hline
\end{tabular}
\end{table}

\begin{figure}[t]
    \centering
    \begin{subfigure}[b]{0.49\linewidth}
        \centering
        \includegraphics[width=\linewidth]{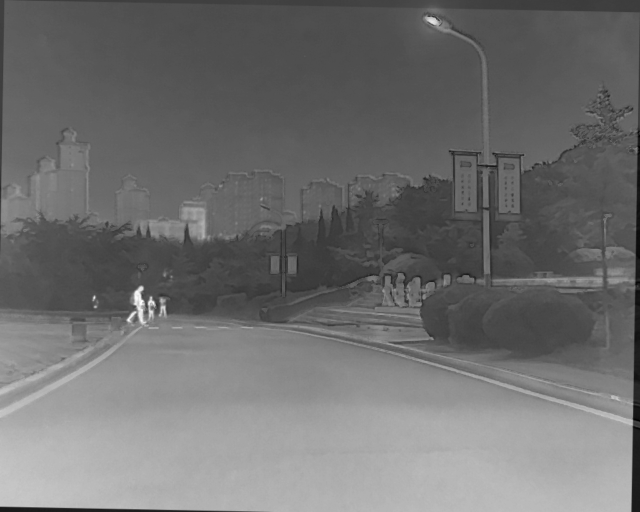}
        \caption{}
        \label{fig:night1}
    \end{subfigure}
    \hfill
    \begin{subfigure}[b]{0.49\linewidth}
        \centering
        \includegraphics[width=\linewidth]{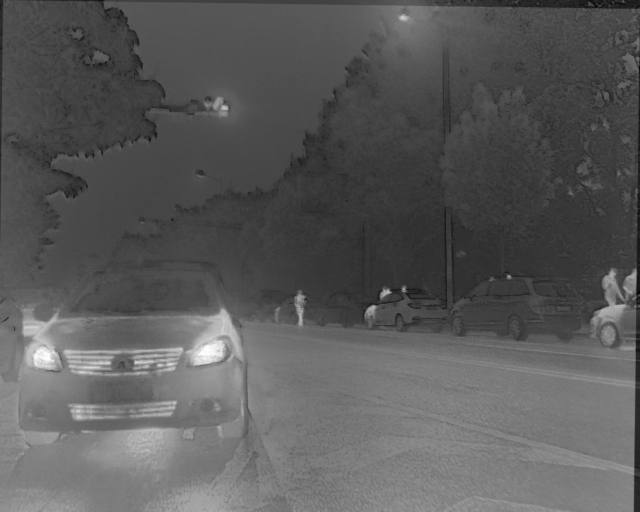}
        \caption{}
        \label{fig:night2}
    \end{subfigure}
    \caption{Fusion results at night. (a) Distant pedestrian and road structure. (b) Multiple pedestrians and vehicles with preserved details.}
    \label{fig:night_results}
\end{figure}

\subsection{Qualitative Results}
In addition to numerical evaluation, we provide qualitative examples to illustrate the interpretability and effectiveness
of FusionNet. As shown in Figure~\ref{fig:fusion_result}, the
fused images preserve thermal targets such as pedestrians and vehicles while retaining fine-grained visual textures. The corresponding alpha maps further demonstrate the adaptive blending behavior of the model, where IR regions dominate for heat sources while VIS regions contribute to structural details.

Moreover, as illustrated in Figure~\ref{fig:night_results}, our method performs effectively in challenging night scenes. The fused images clearly highlight pedestrians, ensuring visibility of critical thermal targets, while simultaneously preserving crosswalks, road structures, and distant buildings. This demonstrates FusionNet’s ability to enhance both safety-relevant objects and broader scene context under low-light conditions.

\section{Conclusion}

In this work, we presented \textbf{FusionNet}, a novel end-to-end framework for infrared and visible image fusion. FusionNet combines a modality-aware attention mechanism, a pixel-wise alpha blending 
module, and a target-aware loss function to achieve adaptive and interpretable fusion. By leveraging weak ROI supervision, the network effectively preserves semantic consistency in task-critical regions, while maintaining structural fidelity and visual detail.

Experiments on the M3FD dataset validate that FusionNet generates fused images with enhanced perceptual quality and clear interpretability \cite{liu2022target}. 
The learned alpha maps provide explicit insights into how different modalities contribute at the pixel level, making the framework both transparent and practical. 

Overall, FusionNet offers a general and extensible solution for semantic-aware multi-modal fusion. In future work, this framework can be extended to other tasks such as multi-sensor perception, video-level fusion, and domain adaptation for robust scene 
understanding in complex environments.

\bibliographystyle{IEEEtran}  
\bibliography{references}   

\begin{thebibliography}{10}
\providecommand{\url}[1]{#1}
\csname url@samestyle\endcsname
\providecommand{\newblock}{\relax}
\providecommand{\bibinfo}[2]{#2}
\providecommand{\BIBentrySTDinterwordspacing}{\spaceskip=0pt\relax}
\providecommand{\BIBentryALTinterwordstretchfactor}{4}
\providecommand{\BIBentryALTinterwordspacing}{\spaceskip=\fontdimen2\font plus
\BIBentryALTinterwordstretchfactor\fontdimen3\font minus \fontdimen4\font\relax}
\providecommand{\BIBforeignlanguage}[2]{{%
\expandafter\ifx\csname l@#1\endcsname\relax
\typeout{** WARNING: IEEEtran.bst: No hyphenation pattern has been}%
\typeout{** loaded for the language `#1'. Using the pattern for}%
\typeout{** the default language instead.}%
\else
\language=\csname l@#1\endcsname
\fi
#2}}
\providecommand{\BIBdecl}{\relax}
\BIBdecl

\bibitem{bellavia2024image}
F.~Bellavia, Z.~Zhao, L.~Morelli, and F.~Remondino, ``Image matching filtering and refinement by planes and beyond,'' \emph{arXiv preprint arXiv:2411.09484}, 2024.

\bibitem{10986975}
D.~Yu, L.~Liu, S.~Wu, K.~Li, C.~Wang, J.~Xie, R.~Chang, Y.~Wang, Z.~Wang, and R.~Ji, ``Machine learning optimizes the efficiency of picking and packing in automated warehouse robot systems,'' in \emph{2025 IEEE International Conference on Electronics, Energy Systems and Power Engineering (EESPE)}, 2025, pp. 1325--1332.

\bibitem{ding2025nerf}
T.~K. Ding, D.~Xiang, Y.~Qi, Z.~Yang, Z.~Zhao, T.~Sun, P.~Feng, and H.~Wang, ``Nerf-based defect detection,'' in \emph{International Conference on Remote Sensing, Mapping, and Image Processing (RSMIP 2025)}, vol. 13650.\hskip 1em plus 0.5em minus 0.4em\relax SPIE, 2025, pp. 368--373.

\bibitem{ding2025neural}
T.~Ding, D.~Xiang, P.~Rivas, and L.~Dong, ``Neural pruning for 3d scene reconstruction: Efficient nerf acceleration,'' \emph{arXiv preprint arXiv:2504.00950}, 2025.

\bibitem{liu2022target}
J.~Liu, X.~Fan, Z.~Huang, G.~Wu, R.~Liu, W.~Zhong, and Z.~Luo, ``Target-aware dual adversarial learning and a multi-scenario multi-modality benchmark to fuse infrared and visible for object detection,'' in \emph{Proceedings of the IEEE/CVF conference on computer vision and pattern recognition}, 2022, pp. 5802--5811.

\bibitem{wu2025warehouse}
S.~Wu, L.~Fu, R.~Chang, Y.~Wei, Y.~Zhang, Z.~Wang, L.~Liu, H.~Zhao, and K.~Li, ``Warehouse robot task scheduling based on reinforcement learning to maximize operational efficiency,'' \emph{Authorea Preprints}, 2025.

\bibitem{zhang2025conditional}
H.~Zhang, H.~Xu, H.~Liu, X.~Yu, X.~Zhang, and C.~Wu, ``Conditional variational underwater image enhancement with kernel decomposition and adaptive hybrid normalization,'' \emph{Neurocomputing}, p. 130845, 2025.

\bibitem{zhou2024reconstruction}
Y.~Zhou, Z.~Zeng, A.~Chen, X.~Zhou, H.~Ni, S.~Zhang, P.~Li, L.~Liu, M.~Zheng, and X.~Chen, ``Evaluating modern approaches in 3d scene reconstruction: Nerf vs gaussian-based methods,'' in \emph{2024 6th International Conference on Data-driven Optimization of Complex Systems (DOCS)}.\hskip 1em plus 0.5em minus 0.4em\relax IEEE, 2024, pp. 926--931.

\bibitem{xu2025robustanomalydetectionnetwork}
\BIBentryALTinterwordspacing
Z.~Xu and Y.~Liu, ``Robust anomaly detection in network traffic: Evaluating machine learning models on cicids2017,'' 2025. [Online]. Available: \url{https://arxiv.org/abs/2506.19877}
\BIBentrySTDinterwordspacing

\bibitem{li2018densefuse}
H.~Li and X.-J. Wu, ``Densefuse: A fusion approach to infrared and visible images,'' \emph{IEEE Transactions on Image Processing}, vol.~28, no.~5, pp. 2614--2623, 2018.

\bibitem{liu2024infrared}
J.~Liu, G.~Wu, Z.~Liu, D.~Wang, Z.~Jiang, L.~Ma, W.~Zhong, and X.~Fan, ``Infrared and visible image fusion: From data compatibility to task adaption,'' \emph{IEEE Transactions on Pattern Analysis and Machine Intelligence}, 2024.

\bibitem{fu2019dual}
J.~Fu, J.~Liu, H.~Tian, Y.~Li, Y.~Bao, Z.~Fang, and H.~Lu, ``Dual attention network for scene segmentation,'' in \emph{Proceedings of the IEEE/CVF conference on computer vision and pattern recognition}, 2019, pp. 3146--3154.

\bibitem{zang2023texture}
S.~Zang, M.~Chen, Z.~Ai, J.~Chi, G.~Yang, C.~Chen, and T.~Yu, ``Texture-aware gray-scale image colorization using a bistream generative adversarial network with multi scale attention structure,'' \emph{Engineering Applications of Artificial Intelligence}, vol. 122, p. 106094, 2023.

\bibitem{liu2025gated}
S.~Liu, Y.~Zhang, X.~Li, Y.~Liu, C.~Feng, and H.~Yang, ``Gated multimodal graph learning for personalized recommendation,'' \emph{INNO-PRESS: Journal of Emerging Applied AI}, vol.~1, no.~1, 2025.

\bibitem{huang_ai-augmented_2025}
\BIBentryALTinterwordspacing
S.~Huang, Y.~Kang, G.~Shen, and Y.~Song, ``{AI}-{Augmented} {Context}-{Aware} {Generative} {Pipelines} for {3D} {Content},'' \emph{Preprints}, Aug. 2025, publisher: Preprints. [Online]. Available: \url{https://doi.org/10.20944/preprints202508.0195.v1}
\BIBentrySTDinterwordspacing

\bibitem{wen2025optimizationbidirectionalgatedloop}
\BIBentryALTinterwordspacing
Z.~Wen, R.~Zhang, and C.~Wang, ``Optimization of bi-directional gated loop cell based on multi-head attention mechanism for ssd health state classification model,'' 2025. [Online]. Available: \url{https://arxiv.org/abs/2506.14830}
\BIBentrySTDinterwordspacing

\bibitem{lu2025predicting}
B.~Lu, Z.~Lu, Y.~Qi, H.~Guo, T.~Sun, and Z.~Zhao, ``Predicting asphalt pavement friction by using a texture-based image indicator,'' \emph{Lubricants}, vol.~13, no.~8, p. 341, 2025.

\bibitem{10823054}
K.~Li, L.~Liu, J.~Chen, D.~Yu, X.~Zhou, M.~Li, C.~Wang, and Z.~Li, ``Research on reinforcement learning based warehouse robot navigation algorithm in complex warehouse layout,'' in \emph{2024 6th International Conference on Artificial Intelligence and Computer Applications (ICAICA)}, 2024, pp. 296--301.

\bibitem{liu2023regformer}
J.~Liu, G.~Wang, Z.~Liu, C.~Jiang, M.~Pollefeys, and H.~Wang, ``Regformer: an efficient projection-aware transformer network for large-scale point cloud registration,'' in \emph{Proceedings of the IEEE/CVF International Conference on Computer Vision}, 2023, pp. 8451--8460.

\bibitem{li2025bideeplab}
J.~Li and Y.~Zhou, ``Bideeplab: An improved lightweight multi-scale feature fusion deeplab algorithm for facial recognition on mobile devices,'' \emph{Computer Simulation in Application}, vol.~3, no.~1, pp. 57--65, 2025.

\bibitem{zhang2023improving}
T.~Zhang, K.~Kasichainula, D.-W. Jee, I.~Yeo, Y.~Zhuo, B.~Li, J.-s. Seo, and Y.~Cao, ``Improving the efficiency of cmos image sensors through in-sensor selective attention,'' in \emph{2023 IEEE International Symposium on Circuits and Systems (ISCAS)}.\hskip 1em plus 0.5em minus 0.4em\relax IEEE, 2023, pp. 1--4.

\bibitem{zhang2024transformer}
T.~Zhang, K.~Kasichainula, Y.~Zhuo, B.~Li, J.-S. Seo, and Y.~Cao, ``Transformer-based selective super-resolution for efficient image refinement,'' in \emph{Proceedings of the AAAI Conference on Artificial Intelligence}, vol.~38, no.~7, 2024, pp. 7305--7313.

\bibitem{liang2025search}
Y.~Liang, D.~Xiang, and X.~Li, ``Search: A self-evolving framework for network architecture optimization,'' \emph{Neurocomputing}, p. 130980, 2025.

\bibitem{wangbmvc}
\BIBentryALTinterwordspacing
F.~Wang, A.~Nayak, Y.~Agrawal, and R.~Shilkrot, ``Hierarchical image link selection scheme for duplicate structure disambiguation,'' in \emph{British Machine Vision Conference 2018, {BMVC} 2018, Newcastle, UK, September 3-6, 2018}.\hskip 1em plus 0.5em minus 0.4em\relax {BMVA} Press, 2018, p. 221. [Online]. Available: \url{http://bmvc2018.org/contents/papers/0718.pdf}
\BIBentrySTDinterwordspacing

\bibitem{niu2025decoding}
T.~Niu, T.~Liu, Y.~T. Luo, P.~C.-I. Pang, S.~Huang, and A.~Xiang, ``Decoding student cognitive abilities: a comparative study of explainable ai algorithms in educational data mining,'' \emph{Scientific Reports}, vol.~15, no.~1, p. 26862, 2025.

\bibitem{wang2021topotxr}
F.~Wang, S.~Kapse, S.~Liu, P.~Prasanna, and C.~Chen, ``Topotxr: A topological biomarker for predicting treatment response in breast cancer,'' in \emph{Information Processing in Medical Imaging: 27th International Conference, IPMI 2021, Virtual Event, June 28--June 30, 2021, Proceedings}.\hskip 1em plus 0.5em minus 0.4em\relax Springer, 2021, pp. 386--397.

\bibitem{zhong2023fusion}
R.~Zhong, Y.~Fu, Y.~Song, and C.~Han, ``A fusion approach to infrared and visible images with gabor filter and sigmoid function,'' \emph{Infrared Physics \& Technology}, vol. 131, p. 104696, 2023.

\bibitem{zhao2024balf}
Z.~Zhao, ``Balf: Simple and efficient blur aware local feature detector,'' in \emph{Proceedings of the IEEE/CVF Winter Conference on Applications of Computer Vision}, 2024, pp. 3362--3372.

\bibitem{huang2024ar}
S.~Huang, Y.~Song, Y.~Kang, C.~Yu \emph{et~al.}, ``Ar overlay: Training image pose estimation on curved surface in a synthetic way,'' in \emph{CS \& IT Conference Proceedings}, vol.~14, no.~17.\hskip 1em plus 0.5em minus 0.4em\relax CS \& IT Conference Proceedings, 2024.

\bibitem{ma2017infrared}
J.~Ma, Z.~Zhou, B.~Wang, and H.~Zong, ``Infrared and visible image fusion based on visual saliency map and weighted least square optimization,'' \emph{Infrared Physics \& Technology}, vol.~82, pp. 8--17, 2017.

\bibitem{zhao2023benchmark}
Z.~Zhao and B.~M. Chen, ``Benchmark for evaluating initialization of visual-inertial odometry,'' in \emph{2023 42nd Chinese Control Conference (CCC)}.\hskip 1em plus 0.5em minus 0.4em\relax IEEE, 2023, pp. 3935--3940.

\bibitem{wang2004image}
Z.~Wang, A.~C. Bovik, H.~R. Sheikh, and E.~P. Simoncelli, ``Image quality assessment: from error visibility to structural similarity,'' \emph{IEEE transactions on image processing}, vol.~13, no.~4, pp. 600--612, 2004.

\bibitem{wang2025evaluating}
C.~Wang, C.~Nie, and Y.~Liu, ``Evaluating supervised learning models for fraud detection: A comparative study of classical and deep architectures on imbalanced transaction data,'' \emph{arXiv preprint arXiv:2505.22521}, 2025.

\end{thebibliography}
\end{document}